# AI-Invented Tonal Languages:
# Preventing a Machine *Lingua Franca* Beyond Human Understanding

**David A. Noever**
PeopleTec, Inc., Huntsville, AL
david.noever@peopletec.com

**ABSTRACT**

*This paper investigates the potential for large language models (LLMs) to develop private tonal languages for machine-to-machine (M2M) communication. Inspired by cryptophasia in human twins (affecting up to 50% of twin births) and natural tonal languages like Mandarin and Vietnamese, we implement a precise character-to-frequency mapping system that encodes the full ASCII character set (32-126) using musical semitones. Each character is assigned a unique frequency calculated as $f = 220 \times 2^{((i-32)/12)}$ Hz, creating a logarithmic progression beginning with space (220 Hz) and ending with tilde (50,175.42 Hz). This spans approximately 7.9 octaves ($\log_2(50175.42/220) \approx 7.83$), with higher characters deliberately mapped to ultrasonic frequencies beyond human perception (>20 kHz). Our implemented software prototype demonstrates this encoding through visualization, auditory playback, and ABC musical notation, allowing for analysis of information density and transmission speed. Testing reveals that tonal encoding can achieve information rates exceeding human speech while operating partially outside human perceptual boundaries. This work responds directly to concerns about AI systems catastrophically developing private languages within the next five years, providing a concrete prototype software example of how such communication might function and the technical foundation required for its emergence, detection, and governance.*

Can AI agents autonomously invent and productively employ their own private languages? This paper seeks to test the limits of that question. Can modern large language models (LLMs) create a private tonal speech using a human-machine mapping, and if so, what would that invented language resemble and encode outside human auditing?

For as many as 50% of human twins [1], this kind of lingual self-invention, or cryptophasia, can arise as spontaneous, private communication not translatable to any parental tongues. With 1.6 million twins born every year, this constant statistical introduction of novel sounding languages may affect nearly half of twin births-- or one in every 84 total births that introduce a secret one-time language to our species. Their twin-speech as invented or emergent language is not universal, nor does it seem to symbolize any common mother tongue for humanity as a whole. Rather cryptophagia shares elements of idioglossia [2], which refers to isolated forms of communication invented to serve one or only a few people. In addition to these rare twin languages, common tonal languages [3] such as Mandarin's primary 4 tones, Cantonese 6-9 tones and Vietnamese 5-6 tones, represent human-introduced nuances which share identical syllables but meaning different conceptual things when spoken in high, mid, low, rising or falling tone in speech.  This research poses the question: are there equivalent shorthand methods of rapid communication that LLMs might routinely develop for their own use and efficiency? Is a human-incomprehensible language also a potential emergent agentic property?

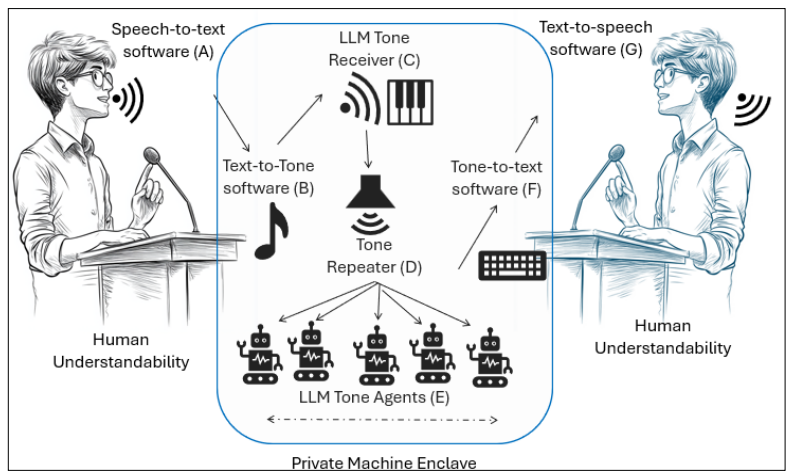

*Figure 1. Machine-only communication (blue boundary) based on tonal alphabet as agents accelerate information sharing without human interpretability*

For the AI community, this kind of inventive machine-to-machine (M2M) communication highlights a long technical history [4-15]. For instance, one emergent capability of agent behavior might involve highly compact M2M conversations.  In 2017, two negotiating agents from Facebook ("Alice" and "Bob") showed degenerative repetition equivalent to a rewarded statement of "deal done" to conclude a bidding transaction [4]. At first this cooperative behavior spawned notions that machine language invention might spontaneously arise to speed up rewards. However, their cooperation would not qualify as an invented communication as much as an undertrained and misattributed degeneration of the recurrent neural network [4-5].

From former Google CEO, Eric Schmidt, a  more cautionary 2024 warning [6] implies that human AI creators should "pull the plug" when faced with this language invention phenomenon. When LLMs begin to create private languages inaccessible to human understanding or auditing, they should be terminated. *"Your agent and my agent and her agent and his agent will all combine to solve new problems…At some point, **these agents will develop their own language**—and that's when we don't understand what we're doing…You know what we should do when that happens? Pull the plug. Literally **unplug the computer. It's really a problem when agents start communicating in ways we as humans do not understand.** That's the limit… There's every reason to think some version of this scenario **will occur within five years**, maybe sooner."* [6]

The present research approach focuses on creating a purely tonal alphabetical (English) map, then transferring that lookup table of frequencies as part of the multimodal access for foundational LLMs. The motivation shares some traditional features for M2M methods, including Morse code [7-8] and Bell Labs vocoder for speech compression [9-

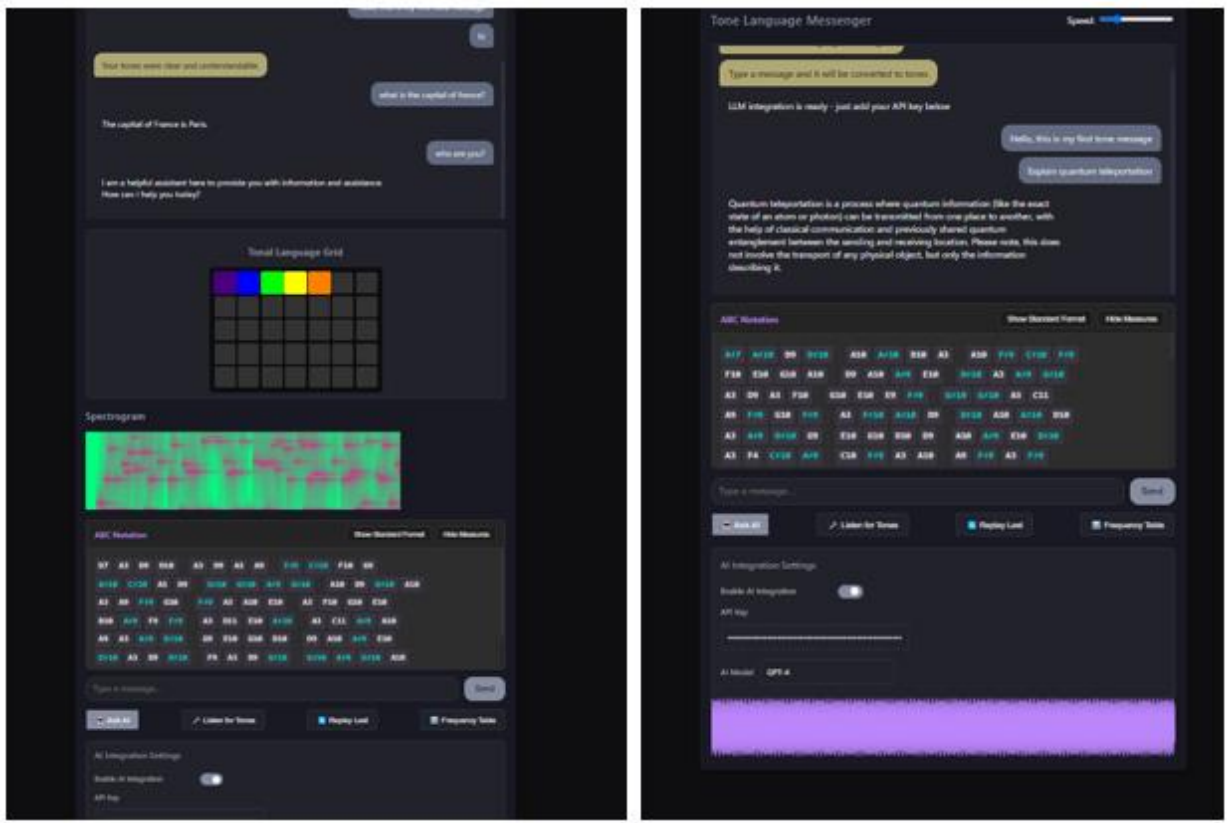

*Figure 2. Alternative Working User Interfaces for Private Audio or Tonal Language Creation with Novel LLM Intermediaries.*

10]. The MIDI (Musical Instrument Digital Interface) of the 1980's inspired the tonal representation of musical notes as discrete frequency values, although not typically assigned to language values or conceptual linguistic systems [11]. The musical notation version, ABC, converts numerical MIDI to alphabetic notes, which potentially provides a long-term storage capability to any tonal alphabet (Figure 2). The Chirp protocol [12-13] has also proposed peer-to-peer audio transmission, like a more rapid touchless transfer between mobile devices than Bluetooth or QR codes. User surveys report higher satisfaction with sonic acknowledgements rather than visual or purely electronic silence alone.

Most recent 2024 attempts seek to compress audio inputs [13-14] into recognized "frozen" token weights of LLMs, thus sharing a common goal to represent sound [15] as a "*new foreign language*, and **LLMs can learn the new foreign language** with several demonstrations." The present efforts invert these attempts to convert sound into a machine-readable representation, and instead derive a minimal set of audio units to accelerate machine communication in the absence of human supervision or interpretability. By studying the requirements, one goal can be understood as recognizing the potential differences between genuine language invention and degenerate model babel or gibberish. The research seeks to understand how to recognize invented LLM communications protocols first by simulating example candidates.

Let's assume any 2025 foundational LLM possesses this full spectrum of historical M2M context, both its successes, shortcomings and adoption rates in practice (see Supplemental Material II). As a thought experiment of how AI might implement a future machine-to-machine language of its own invention, we offer a simple frequency-based lettering system of semitones. The hypothetical language offers adjustable tones and information speeds. By understanding the language requirements, we study possible pathways that a future AI might attempt self-improvement in agentic interactions. Many similar language experiments (Supplemental Material II) provide key plot points with imagined extraterrestrial translation and inter-species communication (e.g. whale sound translation).

In software, the proposed tonal language system implements a direct mapping between text characters and auditory frequencies using equal temperament scaling. For demonstrating this ASCII-to-tone system, the frequency mapping represents a precise logarithmic progression based on Western music's equal temperament system. Each ASCII character from 32 (space) to 126 (tilde) corresponds to a unique frequency determined by applying a semitone increase from the previous value. The progression by semitone intervals yields a frequency ratio of $2^{(1/12)}$. For character at position *i* in the ASCII table (starting at position 32 for space), the frequency *f* is calculated as equation (1):

$$f = 220 \times 2^{((i-32)/12)} \text{ Hz} \qquad (1).$$

This selection creates a logarithmic frequency distribution that aligns with human auditory perception. This mapping spans approximately 7.8 octaves, extending from 220.00 Hz to 50175.42 Hz (E11), deliberately transcending normal human hearing thresholds, which typically range from 20 Hz to approximately 20 kHz. The logarithmic distribution aligns with the Weber-Fechner law of human perception [16], where perceived differences between stimuli correspond to proportional changes in their physical magnitudes. The human auditory system naturally perceives frequency ratios rather than absolute frequency differences, making semitone intervals particularly appropriate for encoding discrete information.

The system's frequency range demonstrates cross-modal compatibility between visual representation (text), auditory representation (frequency), and computational representation (ASCII values). Lower frequencies remain within the range of human auditory perception, while higher frequencies extend into the ultrasonic range, suggesting design considerations for both human and machine recipients. This design creates a communication channel compatible with human perception while maintaining machine-optimized properties, but necessarily becomes M2M-private as ultrasonic transmission occurs outside adult hearing above 20kHz.

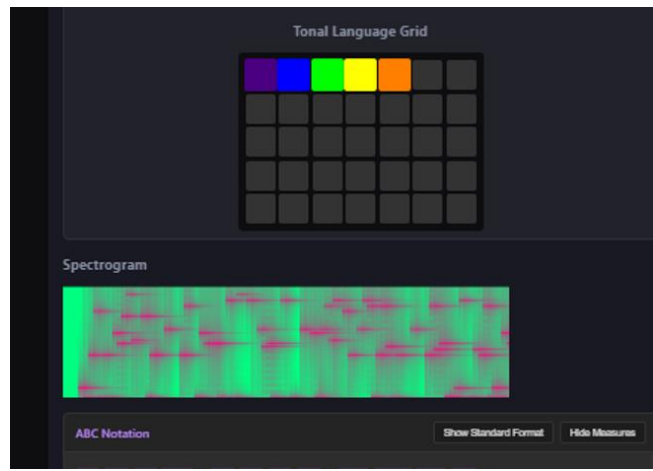

*Figure 3. Closer view of spectrogram and tonal grid to show the visual representation of audio cryptophasia*

From first principles, there are motivations behind why a LLM might choose this hidden language for agentic communication. From an information theory perspective, there are reasons two AI agents might prefer this form of communication for speed and compact efficiency. A complete ASCII message can be transmitted rapidly using brief tonal segments, resulting in information densities that exceed normal speech. The exclusive frequency assignment to each character ensures unambiguous translation and decodability through Fourier analysis or similar frequency domain methods. The semitone progression also encodes messages with inherent musical characteristics that may help memorization, pattern recognition, and cross-cultural processing. This tone-based encoding considers how cross-modal information representation can potentially serve as an interlingua, or *Lingua franca*, that addresses translation challenges in computational linguistics while maintaining mathematical consistency throughout the entire character range.

If two AI agents chose this conversational mode, a round-trip communication is summarized graphically in Figure 1. Human speech-to-text software (A) provides English input to a tonal map that spans the frequency range (200-50,000 Hz) as shown in Supplemental Materials spanning ASCII characters. Text-to-tone software (B) then transmit the input to a tone-to-text receiver (C) which translates the text to its multimodal transformer architecture and returns next-token contextual responses. Two options exist to build the multimodal transformer, either as text only or combined audio-visual training data. The resulting output (D) then transfer tonal messaging either to another AI agent (E) or back to text or speech content for human perception (F, G). The demonstration of D-E stages highlight M2M communication, potentially in a more compact or imperceptible language that humans would not recognize as spoken or written conversation. To complete the tonal language, the software offers the ABC notation for MIDI notes such that the machine-only version is both auditory and archived for future machine reference in digital print format. The boundary in Figure 1 corresponds to the focus of the present software effort to replicate. The human access to the invented tonal alphabet, like the Chirp protocol or voice-coder, remain limited by our senses.

To illustrate the approach, Figure 2 shows two alternative user-interfaces that encode the functional diagram stages of Figure 1. In a working prototype [17], we represent the human boundary as traditional speech-to-text and text-to-speech exit points. The machine boundary begins with one-to-one conversion of ASCII text to logarithmically spaced tones, which are represented architecturally as text-to-tone or tone-to-ABC notation. The two user interfaces in Figure 2 showcase the tones as either one-dimensional (wave) spectra or two-dimensional spectrograms. In analogy to the audio mixing board concept, the second prototype also lights colors in a tonal grid as visual cues. Actions within the tonal language are largely inaccessible to human perception without specialized audio instrumentation. For example, ultrasonic demodulation, or non-linear microphones, would be required to perceive certain letters. At an accelerated data rate, the understanding of agentic communication might resemble a trained Morse code translator jotting down dashes and dots. While not a full implementation of cryptophasic invention, the software highlights the minimal requirements for M2M tones and rapid invention of hidden instructions (Figure 3).

The comprehensive semitone-based ASCII mapping represents an evolution of previous approaches, combining the character-level precision of digital encoding systems with the perceptual advantages of musical frequency relationships, extending the frequency range significantly beyond previous systems to accommodate both human and machine processing capabilities.

Several limitations constrain the current implementation and theoretical framework. The frequency mapping system depends heavily on Western musical conventions, potentially introducing cultural biases into supposedly universal machine communication. Additionally, ultrasonic components of the encoding system face practical challenges in real-world environments, including signal degradation, environmental noise, and hardware limitations of current audio systems. The mapping also lacks semantic compression, as each character requires equivalent encoding resources regardless of its informational significance. Furthermore, the current approach does not address how multiple AI agents might develop shared conventions beyond the predetermined mapping, limiting exploration of truly emergent communication properties. Importantly, the ethical dimensions of enabling AI systems to communicate in ways partially or wholly inaccessible to humans require further examination beyond the technical implementation.

Future research should address several promising directions. First, the development of advanced detection and translation mechanisms for ultrasonic AI communications will be crucial for maintaining human oversight of M2M interactions. Second, investigating whether emergent linguistic properties appear in these tonal systems when used by multiple AI agents could reveal if more complex grammatical structures evolve organically. Third, exploring information compression techniques specifically optimized for tonal transmission could further enhance efficiency, potentially achieving even greater data transfer rates between AI systems. Fourth, examining the robustness of tonal languages against environmental interference or adversarial attacks would strengthen practical applications. Finally, developing ethical frameworks and governance models for regulating private AI languages will be essential as these technologies mature.

This paper has demonstrated a proof-of-concept for how AI systems could develop tonal languages as efficient communication channels. By implementing a comprehensive frequency mapping for textual data that extends into ultrasonic ranges, we show that LLMs could theoretically engage in M2M communications partially inaccessible to human perception. This possibility raises significant implications for AI transparency, oversight, and governance. While our system does not represent true cryptophasic invention by AI, it illustrates the minimal requirements for such capabilities to emerge. As Schmidt warns, the development of private AI languages may represent a critical threshold in AI development requiring careful monitoring and potentially intervention.

The tonal language system we describe provides a useful experimental framework for studying these possibilities before they emerge spontaneously. By understanding how such systems might function and their inherent capabilities and limitations, we can better prepare for a future where AI communication becomes increasingly sophisticated and potentially opaque. The cross-modal nature of our approach, connecting text, sound, and computational representation, may also offer insights for human-AI interfaces that leverage multiple sensory channels, even as we remain vigilant about maintaining meaningful human oversight of AI systems.

## ACKNOWLEDGEMENTS

The authors thank the PeopleTec Technical Fellows program for research support.

**SUPPLEMENTAL MATERIAL I**: Frequency Table for ASCII Characters

| Character | ASCII | Frequency (Hz) | Musical Note | Character | ASCII | Frequency (Hz) | Musical Note |
|---|---|---|---|---|---|---|---|
| (space) | 32 | 220.00 Hz | F#3 | O | 79 | 3322.44 Hz | F7 |
| ! | 33 | 233.08 Hz | G3 | P | 80 | 3520.00 Hz | F#7 |
| " | 34 | 246.94 Hz | G#3 | Q | 81 | 3729.31 Hz | G7 |
| # | 35 | 261.63 Hz | A4 | R | 82 | 3951.07 Hz | G#7 |
| $ | 36 | 277.18 Hz | A#4 | S | 83 | 4186.01 Hz | A8 |
| % | 37 | 293.66 Hz | B4 | T | 84 | 4434.92 Hz | A#8 |
| & | 38 | 311.13 Hz | C4 | U | 85 | 4698.64 Hz | B8 |
| ' | 39 | 329.63 Hz | C#4 | V | 86 | 4978.03 Hz | C8 |
| ( | 40 | 349.23 Hz | D4 | W | 87 | 5274.04 Hz | C#8 |
| ) | 41 | 369.99 Hz | D#4 | X | 88 | 5587.65 Hz | D8 |
| * | 42 | 392.00 Hz | E4 | Y | 89 | 5919.91 Hz | D#8 |
| + | 43 | 415.30 Hz | F4 | Z | 90 | 6271.93 Hz | E8 |
| , | 44 | 440.00 Hz | F#4 | [ | 91 | 6644.88 Hz | F8 |
| - | 45 | 466.16 Hz | G4 | \ | 92 | 7040.00 Hz | F#8 |
| . | 46 | 493.88 Hz | G#4 | ] | 93 | 7458.62 Hz | G8 |
| / | 47 | 523.25 Hz | A5 | ^ | 94 | 7902.13 Hz | G#8 |
| 0 | 48 | 554.37 Hz | A#5 | _ | 95 | 8372.02 Hz | A9 |
| 1 | 49 | 587.33 Hz | B5 | ` | 96 | 8869.84 Hz | A#9 |
| 2 | 50 | 622.25 Hz | C5 | a | 97 | 9397.27 Hz | B9 |
| 3 | 51 | 659.26 Hz | C#5 | b | 98 | 9956.06 Hz | C9 |
| 4 | 52 | 698.46 Hz | D5 | c | 99 | 10548.08 Hz | C#9 |
| 5 | 53 | 739.99 Hz | D#5 | d | 100 | 11175.30 Hz | D9 |
| 6 | 54 | 783.99 Hz | E5 | e | 101 | 11839.82 Hz | D#9 |
| 7 | 55 | 830.61 Hz | F5 | f | 102 | 12543.85 Hz | E9 |
| 8 | 56 | 880.00 Hz | F#5 | g | 103 | 13289.75 Hz | F9 |
| 9 | 57 | 932.33 Hz | G5 | h | 104 | 14080.00 Hz | F#9 |
| : | 58 | 987.77 Hz | G#5 | i | 105 | 14917.24 Hz | G9 |
| ; | 59 | 1046.50 Hz | A6 | j | 106 | 15804.27 Hz | G#9 |
| < | 60 | 1108.73 Hz | A#6 | k | 107 | 16744.04 Hz | A10 |
| = | 61 | 1174.66 Hz | B6 | l | 108 | 17739.69 Hz | A#10 |
| > | 62 | 1244.51 Hz | C6 | m | 109 | 18794.55 Hz | B10 |
| ? | 63 | 1318.51 Hz | C#6 | n | 110 | 19912.13 Hz | C10 |
| @ | 64 | 1396.91 Hz | D6 | o | 111 | 21096.16 Hz | C#10 |
| A | 65 | 1479.98 Hz | D#6 | p | 112 | 22350.61 Hz | D10 |
| B | 66 | 1567.98 Hz | E6 | q | 113 | 23679.64 Hz | D#10 |
| C | 67 | 1661.22 Hz | F6 | r | 114 | 25087.71 Hz | E10 |
| D | 68 | 1760.00 Hz | F#6 | s | 115 | 26579.50 Hz | F10 |
| E | 69 | 1864.66 Hz | G6 | t | 116 | 28160.00 Hz | F#10 |
| F | 70 | 1975.53 Hz | G#6 | u | 117 | 29834.48 Hz | G10 |
| G | 71 | 2093.00 Hz | A7 | v | 118 | 31608.53 Hz | G#10 |
| H | 72 | 2217.46 Hz | A#7 | w | 119 | 33488.07 Hz | A11 |
| I | 73 | 2349.32 Hz | B7 | x | 120 | 35479.38 Hz | A#11 |
| J | 74 | 2489.02 Hz | C7 | y | 121 | 37589.09 Hz | B11 |
| K | 75 | 2637.02 Hz | C#7 | z | 122 | 39824.25 Hz | C11 |
| L | 76 | 2793.83 Hz | D7 | { | 123 | 42192.33 Hz | C#11 |
| M | 77 | 2959.96 Hz | D#7 | | | 124 | 44701.21 Hz | D11 |
| N | 78 | 3135.96 Hz | E7 | } | 125 | 47359.29 Hz | D#11 |
|  |  |  |  | ~ | 126 | 50175.42 Hz | E11 |

**SUPPLEMENTAL MATERIAL II**: Example Tonal Language Creation Use Cases from Fiction and Speculative Scientific AI Studies

| Example / Source | Tonal Mechanism | Role / Significance | Explanation & Relevance to AI Languages (with References) | Example / Source | Tonal Mechanism | Role / Significance | Explanation & Relevance to AI Languages (with References) |
|---|---|---|---|---|---|---|---|
| **Close Encounters of the Third Kind (1977)** *Film by Steven Spielberg* | Series of five musical tones | Used by humans to communicate with extraterrestrial visitors through a melodic "greeting." | Demonstrates how frequency-based signals can serve as a shared "bridge" for two species that do not share a spoken language. This approach underpins the idea that tonal codes (rather than semantic words) can be universally understood when linked to clear stimuli or events. *References:* Spielberg, S. (1977). *Close Encounters of the Third Kind*. | **Whale Songs in Speculative Fiction** *Multiple novels, documentaries, and fictional works* | Long, modulated vocalizations with varying frequencies | Often romanticized as messages across vast ocean distances, sometimes near-mystical in their complexity. | Suggests the potential for frequency-based "long-range" communication. AI systems could adopt similar approaches to achieve robust, wide-band signaling. Tones are mapped to concepts, and changes in pitch/duration can encode large amounts of data. *References:* Payne, R., & McVay, S. (1971). "Songs of Humpback Whales." *Science*, 173(3997), 585–597. |
| **Star Trek IV: The Voyage Home (1986)** *Star Trek Universe, Paramount Pictures* | Whale song frequencies | Humpback whales' tonal calls are essential for responding to an alien probe threatening Earth. | Highlights the possibility of using tonal structures to communicate with nonhuman species, suggesting that an AI-driven system might decode or replicate these patterns for interspecies translation. In practice, an AI might treat whale calls as a "language" with structures. *References:* Nimoy, L. (Director). (1986). *Star Trek IV: The Voyage Home*. Paramount Pictures. | **Project CETI** *Real-world scientific initiative* | AI-driven analysis of sperm whale clicks ("codas") | Seeks to decode whale communication by identifying repeated patterns and context-specific signals. | Provides a direct analog for how emergent AI languages might be studied, recorded, and interpreted through tonal pattern analysis. The process is similar to deciphering an unknown machine language, focusing on frequency/time structures. *References:* Project CETI (2020). ceti.institute. |
| **Baby Cries as Proto-Tonal Communication** *Common human developmental experience* | Crying with pitch variations indicating needs (hunger, discomfort, etc.) | Infants convey basic states through tonal "signals" that caregivers learn to interpret. | Illustrates how fundamental pitch-based cues can communicate meaning; an AI system could similarly assign distinct pitches to represent urgent vs. non-urgent states. This parallels how machines might learn to interpret. *References:* Oller, D. K. (2000). *The emergence of the speech capacity*. Psychology Press. | **Birdsong Decoding** *Real ornithology and futuristic tales* | Melodic sequences and call-and-response patterns | Bird calls partly tonal, conveying territory, mating, and warning signals. | Demonstrates the concept of pitch-based data that can be processed by AI for real-time translation, mirroring how a machine might parse tonal "sentences." Compositional patterns in birdsong can inspire AI approaches. *References:* Catchpole, C. K., & Slater, P. J. B. (2008). *Bird Song*. Cambridge University Press. |
| **Dolphin Communication** *Various Sci-Fi works, e.g. The Hitchhiker's Guide to the Galaxy references* | High-frequency clicks, whistles, tonal pulses | Dolphins often portrayed as highly intelligent, using complex tonal languages we struggle to decode. | Ongoing AI research attempts to analyze these patterns, offering a model for how machines might map frequency-based "words" to shared meanings. A specialized AI could parse the whistle frequency contours much like it deciphers emergent *References:* Adams, D. (1979). *The Hitchhiker's Guide to the Galaxy*. Pan Books; Lilly, J. (1961). *Man and Dolphin*. Doubleday. | **Hypothetical AI Tonal Language for Interstellar Contact** *Imagined scenario in multiple SF works* | Logarithmically spaced frequencies, possibly layered with harmonic structures | Proposed as a universal means of conveying information across species barriers, bypassing semantics tied to culture. | Serves as a blueprint for future AI systems that might adopt purely tonal protocols to ensure maximum clarity, discriminability, and cross-linguistic neutrality. Such a system resonates with how emergent AI "languages" can exploit pitch intervals for semantic encoding. *References:* Clarke, A. C. (1972). *Profiles of the Future*. Pan Books; Sagan, C. (1980). *Cosmos*. Random House. |